\documentclass[conference]{IEEEtran}
\IEEEoverridecommandlockouts
\usepackage{cite}
\usepackage{amsmath,amssymb,amsfonts}
\usepackage{algorithmic}
\usepackage{graphicx}
\usepackage{textcomp}
\def\BibTeX{{\rm B\kern-.05em{\sc i\kern-.025em b}\kern-.08em
T\kern-.1667em\lower.7ex\hbox{E}\kern-.125emX}}
\begin{document}

\newcommand{\fourier}[1]{\mathcal{F} \biggl[#1 \biggl]}
\newcommand{\ifourier}[1]{\mathcal{F}^{-1} \biggl[#1 \biggl]}
\newcommand{\sfourier}[1]{\mathcal{F} \bigl[#1 \bigr]}
\newcommand{\sifourier}[1]{\mathcal{F}^{-1} \bigl[#1 \bigr]}
\newcommand{\fft}[1]{FFT \biggl[#1 \biggl]} 
\newcommand{\ifft}[1]{FFT^{-1} \biggl[#1 \biggl]} 
\newcommand{\pd}[2]{ \frac{\partial #1}{\partial #2} }
\newcommand{\absbar}[1]{\left |#1 \right |} 
\newcommand{\dop}[3]{{{\rm D}^{#1}_{#2} [#3]}} 

\newcommand{\prop}[2]{\mathcal{{\rm Prop}}_{#1} \bigl[#2\bigl]}

\newcommand{\eq}[1]{Eq.(\ref{#1})}
\newcommand{\eqs}[2]{Eqs.(\ref{#1})-(\ref{#2})}
\newcommand{\fig}[1]{Fig.\ref{#1}}
\newcommand{\figs}[2]{Figs.\ref{#1}-\ref{#2}}
\newcommand{\Fig}[1]{Figure \ref{#1}}

\title{Convolutional neural network-based regression for depth prediction in digital holography\\
}

\author{\IEEEauthorblockN{Tomoyoshi Shimobaba}
\IEEEauthorblockA{\textit{Graduate School of Engineering} \\
\textit{Chiba University}\\
1-33 Yayoi-cho, Inage-ku, Chiba, Japan \\
shimobaba@faculty.chiba-u.jp}
\and
\IEEEauthorblockN{Takashi Kakue}
\IEEEauthorblockA{\textit{Graduate School of Engineering} \\
\textit{Chiba University}\\
1-33 Yayoi-cho, Inage-ku, Chiba, Japan \\
t-kakue@chiba-u.jp}
\and
\IEEEauthorblockN{Tomoyoshi Ito}
\IEEEauthorblockA{\textit{Graduate School of Engineering} \\
\textit{Chiba University}\\
1-33 Yayoi-cho, Inage-ku, Chiba, Japan \\
itot@faculty.chiba-u.jp}
}

\maketitle

\begin{abstract}
Digital holography enables us to reconstruct objects in three-dimensional space from holograms captured by an imaging device. 
For the reconstruction, we need to know the depth position of the recoded object in advance. 
In this study,  we propose depth prediction using convolutional neural network (CNN)-based regression. 
In the previous researches, the depth of an object was estimated through reconstructed images at different depth positions from a hologram using a certain metric that indicates the most focused depth position; however, such a depth search is time-consuming. The CNN of the proposed method can directly predict the depth position with millimeter precision from holograms.
\end{abstract}


\section{Introduction}
Digital holography is a promising imaging technique because it enables simultaneously measuring the amplitude and phase of objects, and reconstructing objects in three-dimensional (3D) space from a hologram captured by an imaging sensor \cite{poon2006digital}. 
Digital holography can be applied from microscopic \cite{kim2010principles} objects (digital holographic microscopy) to macroscopic objects \cite{nakatsuji2008free}.
For measuring the objects in 3D space, we need to know the depth position of the recorded object in advance.
Subsequently diffraction is calculated at that position.
When using the angular spectrum method \cite{goodman2005introduction}, the diffraction calculation is performed by the following equation:
\begin{equation}
u_z(x,y) = \sifourier{\sfourier{u_h(x,y)} H(\mu, \nu)},
\label{eqn:diffraction}
\end{equation}
where $u_h(x,y)$ and $u_z(x,y)$ is a hologram and the complex amplitude in the reconstructed plane at depth $z$, and the operators $\sfourier{\cdot}$ and $\sifourier{\cdot}$ are the Fourier transform and its inverse transform, respectively. $H(\mu, \nu)$ is the transfer function of the angular spectrum method \cite{goodman2005introduction}.

Without knowing the object's depth position in advance, in previous researches, depth search was performed by performing multiple diffraction calculations for different depth parameters $z$, and detecting the most focused depth position by certain focusing metrics.

A number of focusing metrics in digital holography have been proposed; e.g., entropy-based methods \cite{han2010laplacian, ren2015autofocusing, jiao2017enhanced}, a Fourier spectrum-based method \cite{langehanenberg2008autofocusing} and a wavelet-based method \cite{liebling2004autofocus}.
Moreover, the Laplacian ,variance, gradient, and Tamura coefficients of the intensity of a reconstructed image have been used as focused metrics \cite{langehanenberg2008autofocusing, memmolo2011automatic}. 
Almost all the methods measure the sharpness of the reconstructed intensity images and judge the sharpest image as the focused image.

The Tamura coefficient is easier to find global solution of depth position compared to the Laplacian, variance, gradient metrics \cite{memmolo2011automatic}. The coefficient $C_z$ is calculated as
\begin{equation}
C_z = \frac{\sigma_{I_z}}{ \bar{I_z} },
\label{eqn:tamura}
\end{equation}
where $I_z=|u_z(x,y)|^2$ is the reconstructed intensity, $\sigma_I$ and $\bar{I}$ are the standard deviation and average of the intensity, respectively.
In this study, we use the Tamura coefficient for comparison to the proposed method. 
The depth search using this metric is performed as follows: we calculate multiple reconstructed intensities $I_z$ using the diffraction calculation equation \eq{eqn:diffraction} and varying changing the depth position $z$ from $z_1$ to $z_2$ with a step $\delta z$.
Subsequently, we calculate the focused metrics $C_z$ for the reconstructed intensities. 
We can find the focused depth position at the maximum $C_z$.
However, such a depth search is time-consuming, because it requires multiple diffraction calculations.

This study proposes depth prediction using convolutional neural network (CNN)-based regression. 
The CNN in the proposed method can directly predict the depth position with millimeter precision from holograms, without multiple diffraction calculations.

\section{Proposed method}
Recently, deep learning \cite{goodfellow2016deep} has been intensively investigated and applied in a wide range of research fields. 
Deep learning is a type of neural network, but it has deeper layers than conventional neural networks.
Deep learning has been apllied in the holography, for example, classification \cite{shimobaba2017convolutional, muramatsu2017deepholo, kim2017deep}, complex amplitude restoration in digital holographic microscopy \cite{rivenson2017phase}, and noise suppression of holographic reconstructed images\cite{shimobaba2017autoencoder}.
Among the many deep neural networks, CNN demonstrates excellent performance in the field of image processing, which comprises convolutional layers, pooling layer, and fully connected layers.

In this study, a continuous depth value is predicted by inputting hologram information to the CNN of the proposed method; i.e., the CNN needs to solve a multiple regression problem. 
The pioneer work of depth prediction in digital holography was presented by T. Pitk{\"a}aho, A. Manninen and T. J. Naughton \cite{pitkaaho2017performance,pitkaaho2017focus}.
The difference between the proposed method and those applied in previous studies\cite{pitkaaho2017performance,pitkaaho2017focus} is that in these studies, the depth prediction was solved as a classification problem, so that the predicted depth becomes a discrete value. On the other hand, since the proposed method solves the problem as a multiple regression, the predicted depth becomes a continuous value.

\begin{figure}[htbp]
\centering
\fbox{\includegraphics[width=\linewidth]{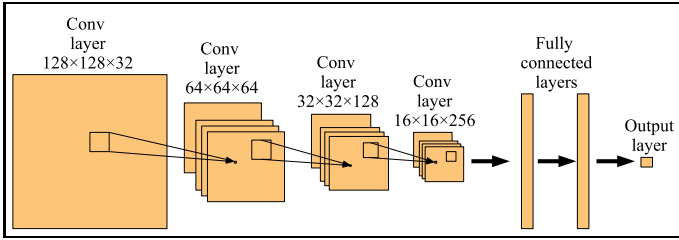}}
\caption{CNN for predicting focused depth.}
\label{fig:cnn}
\end{figure}

\Fig{fig:cnn} depicts the CNN structure in the proposed method.

The input layer is for inputting hologram information.
In this study, we use two types of hologram information, the raw interference pattern and the power spectrum of the hologram, and we will compare the differences in the prediction results later.
The convolution layer performs convolution operations with the kernel size of $3 \times 3$ pixels to acquire feature maps of the input information.
The dimension of the first convolution layer is $128 \times 128 \times 32$ which denotes an input image size of $128 \times 128$ pixels and 32 different convolution kernels.
All the convolution layers are connected to activation functions (ReLU function) and max-pooling layers. 
The dimensions of the second, third, and forth convolution layers are $64 \times 64 \times 64$, $32 \times 32 \times 128$ and $16 \times 16 \times 256$.
The dimension of each fully connected layer is 2,048.
The activation function of the output layer is a linear function (identity function, i.e., $y=x$) because we want to obtain a continuous depth value.

This network is trained by minimizing the loss function, where we use the mean square error (MSE) between the outputs $d^{(j)}_o$ of this network and depth values $d^{(j)}_t$ included in a dataset.
The loss function (MSE) is defined as
\begin{equation}
e = \frac{1}{N} \sum_{j=0}^{N-1} | d^{(j)}_o - d^{(j)}_t |^2,
\label{eqn:mse}
\end{equation}
where the subscript $j$ denotes $j$-th data in the dataset and $N$ is the size of the dataset.
The CNN is trained using Adam optimizer \cite{goodfellow2016deep} with the initial learning rate of 0.0005.
The learning rate is automatically decreased when the MSE is stagnated.

\section{Results}
\begin{figure}[htbp]
\centering
\fbox{\includegraphics[width=\linewidth]{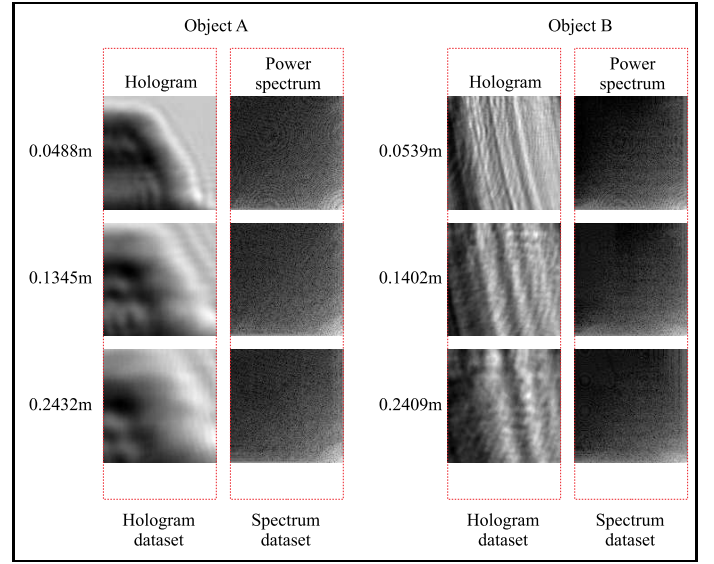}}
\caption{``Hologram dataset'' and ``power spectrum dataset.'' These datasets consist of holograms (or spectra) and the corresponding depth values.}
\label{fig:dataset}
\end{figure}

We prepare two kinds of datasets as illustrated in \fig{fig:dataset}.
The holograms and the power spectra of two objects are presented as examples.
The first dataset consists of raw images of holograms and their depth values.
We use natural images taken from the dataset ``Caltech-256'' \cite{griffin2007caltech} as original objects of the holograms and calculated the holograms as inline hologram from these natural images.
The hologram size is $1,024 \times 1,024$ pixels.
The reference light with the wavelength of 633nm is planar wave.
Twenty holograms were captured while moving a same original object along the depth direction ranging from $0.05$m to $0.25$m at random $\delta z$ intervals of [-0.5mm, 0.5mm]. 

Since the holograms were acquired by an image sensor with $1,024 \times 1,024$ pixels, we extracted the $128 \times 128$ pixels in the center of the hologram.
The reason for reducing the size of the hologram is to speed up the learning and prediction of the CNN.
Accordingly, it helps to simplify the network structure.

The second dataset consists of the power spectra of the holograms and the corresponding depth values, as depicted in \fig{fig:dataset}.
We call the dataset ``spectrum dataset.''
The power spectra are calculated from the holograms of $1,024 \times 1,024$ pixels, and subsequently, the first quadrant of the calculation result is extracted and further reduced to $128 \times 128$ pixels by linear interpolation.

\begin{figure}[htbp]
\centering
\fbox{\includegraphics[width=\linewidth]{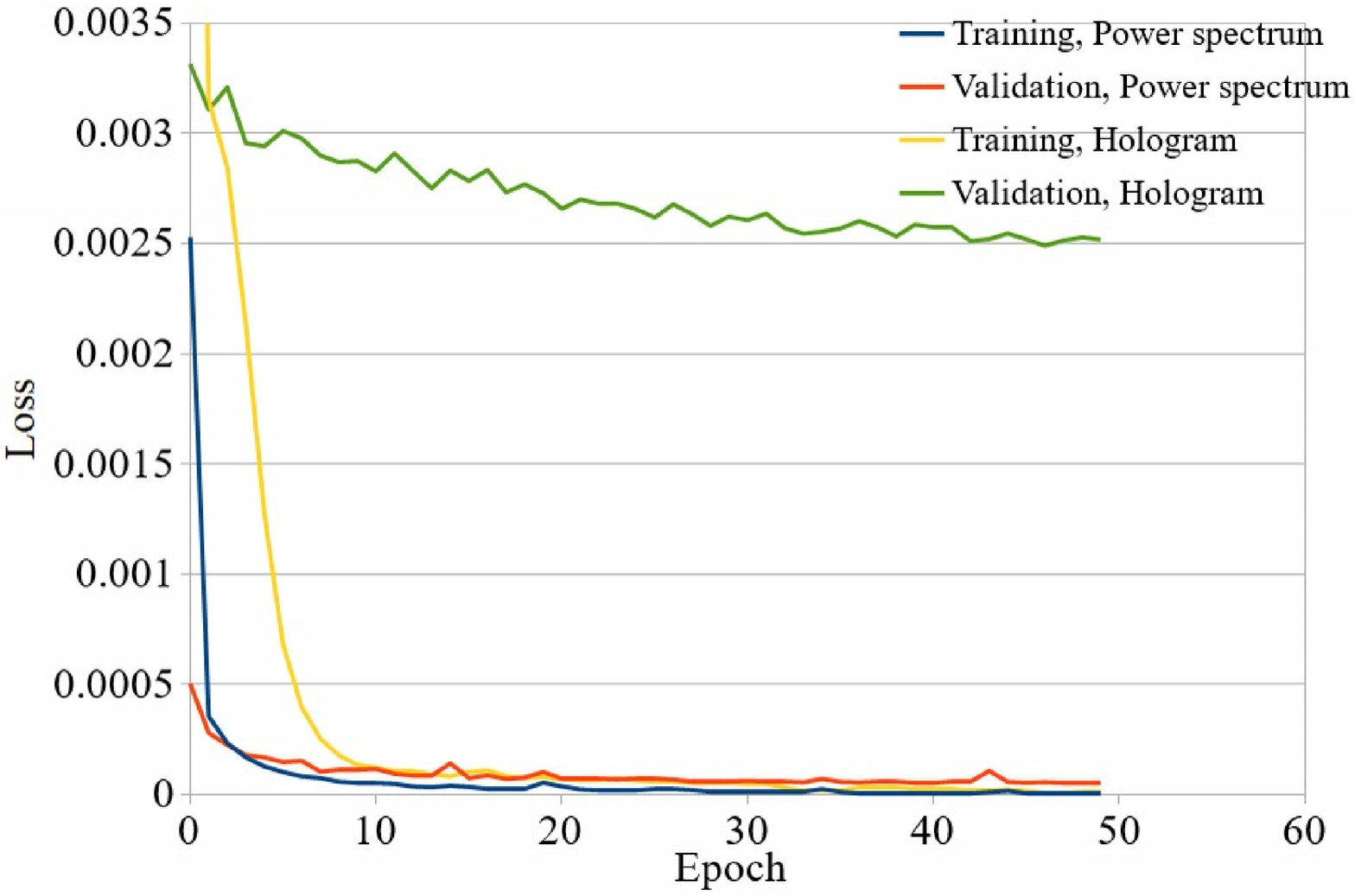}}
\caption{Training and validation losses for the hologram and spectrum datasets with increasing epoch.}
\label{fig:loss}
\end{figure}

The hologram and spectrum datasets are prepared for training and validation, respectively.
\Fig{fig:loss} depicts the change of the loss function for the training and validation datasets with increasing epoch.
As can be seen, in the hologram dataset, the loss value can only be reduced to 0.00249, which means that the average depth error is about 50 mm.
In contrast, the loss value of the spectrum dataset for validation reaches $5.245 \times 10^{-5}$, which means that the average depth error is only 7.2 mm.

\begin{figure}[htbp]
\centering
\fbox{\includegraphics[width=\linewidth]{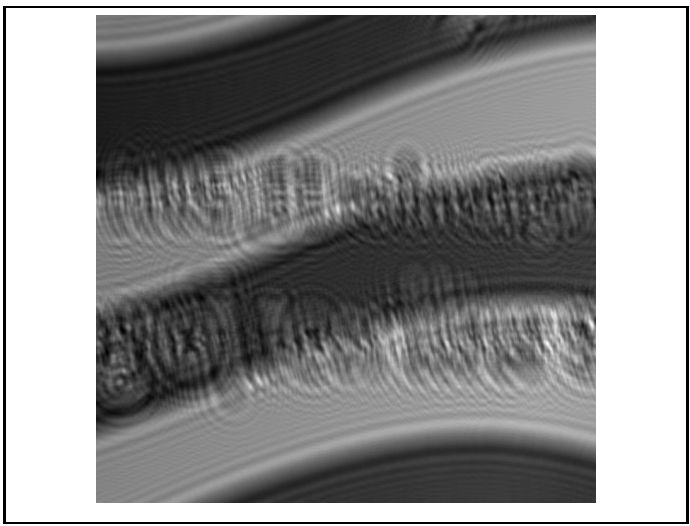}}
\caption{Hologram recorded from an original object at approximately 0.143 m.}
\label{fig:hol}
\end{figure}

\begin{figure}[htbp]
\centering
\fbox{\includegraphics[width=\linewidth]{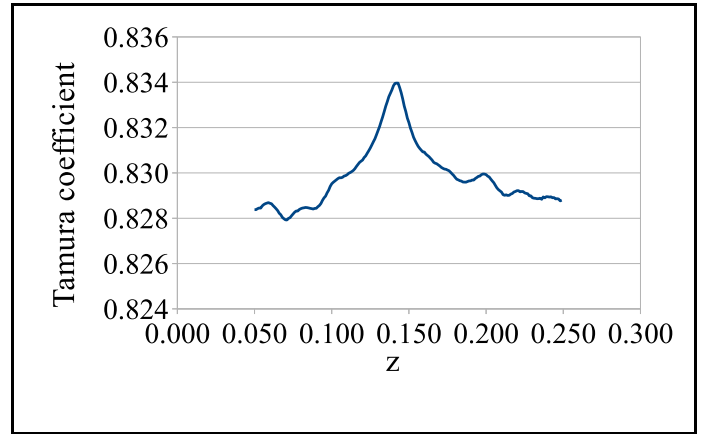}}
\caption{Coefficient plot with increasing the propagation distance of diffraction calculation.}
\label{fig:tamura}
\end{figure}

\begin{figure}[htbp]
\centering
\fbox{\includegraphics[width=\linewidth]{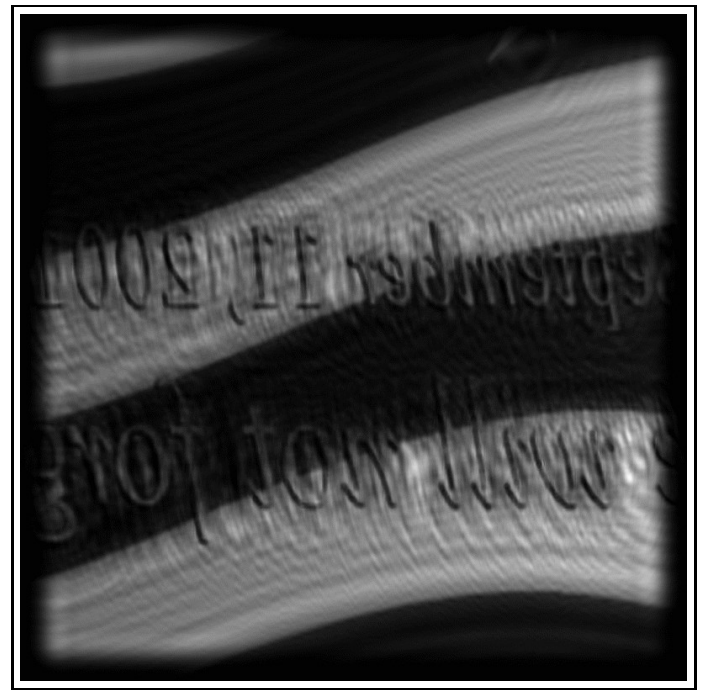}}
\caption{Reconstructed image of the hologram at $z=0.147$ m estimated by the maximum value of Tamura coefficient. The difference between the correct depth and the estimated depth is 4 mm.}
\label{fig:reconst_tamura}
\end{figure}

\Fig{fig:hol} depicts a hologram with $1,024 \times 1,024$ pixels to verify the effectiveness of the CNN.
The hologram is recorded from an original object at 0.143 m .
We perform the depth search using the Tamura coefficient (\eq{eqn:tamura}) ranging from $0.05$ m to $0.25$ m with the depth step of 1 mm.
The coefficient plot is shown in \fig{fig:tamura}.
The plot shows the maximum coefficient at $z=0.147$ m; therefore, in this case, the difference between the correct depth and the calculated depth is 4 mm.
\Fig{fig:reconst_tamura} depicts the reconstructed image of the hologram at $z=0.147$ m.

\begin{figure}[htbp]
\centering
\fbox{\includegraphics[width=\linewidth]{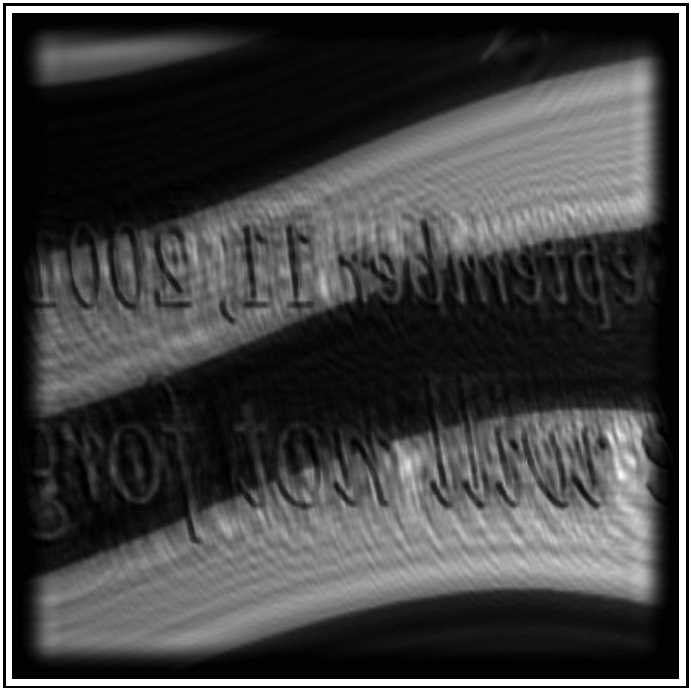}}
\caption{Reconstructed image of the hologram at $z=0.138$m directly predicted by the CNN-based regression. The difference between the correct depth and the predicted depth is 5mm.}
\label{fig:reconst_cnn}
\end{figure}

Subsequently, the CNN can be used to predict the depth value of 0.138 m directly from the power spectrum of the hologram, without the depth search. 
\Fig{fig:reconst_cnn} depicts the reconstructed image of the hologram at $z=0.138$ m.
The difference between the actual depth and the predicted depth is 5 mm.

The calculation time of the CNN is 3.5ms per one power spectrum on a NVIDIA GeForce 970 GTX GPU; in contrast,  the calculation time of the depth search in \fig{fig:tamura} is 2,892ms on the same GPU.
Compared with the depth search, the CNN can greatly speed up the calculation of depth prediction.
All the hologram calculations were performed using our wave optics library, CWO++ \cite{shimobaba2012computational}.

\section{Conclusion}
We proposed a CNN-based regression for depth prediction in digital holography. The CNN enables the direct prediction of the depth value with millimeter precision from the power spectrum of a hologram. 
We used a power spectrum that reduced the size to 1/8$^{th}$ of the original holograms. 
This reduction helped simplify the CNN network structure. 
It will be also useful for predicting depth value for larger holograms. 
In future work, we plan to estimate a depth value with a CNN regression using optically recorded holograms and predict the depth values of holograms of multiple recorded objects at different depth position.

\section*{Acknowledgment}
This work was partially supported by JSPS KAKENHI Grant Number 16K00151.

\bibliographystyle{IEEEtran}
\bibliography{reference}

\begin{thebibliography}{10}
\providecommand{\url}[1]{#1}
\csname url@samestyle\endcsname
\providecommand{\newblock}{\relax}
\providecommand{\bibinfo}[2]{#2}
\providecommand{\BIBentrySTDinterwordspacing}{\spaceskip=0pt\relax}
\providecommand{\BIBentryALTinterwordstretchfactor}{4}
\providecommand{\BIBentryALTinterwordspacing}{\spaceskip=\fontdimen2\font plus
\BIBentryALTinterwordstretchfactor\fontdimen3\font minus
  \fontdimen4\font\relax}
\providecommand{\BIBforeignlanguage}[2]{{%
\expandafter\ifx\csname l@#1\endcsname\relax
\typeout{** WARNING: IEEEtran.bst: No hyphenation pattern has been}%
\typeout{** loaded for the language `#1'. Using the pattern for}%
\typeout{** the default language instead.}%
\else
\language=\csname l@#1\endcsname
\fi
#2}}
\providecommand{\BIBdecl}{\relax}
\BIBdecl

\bibitem{poon2006digital}
T.-C. Poon, \emph{Digital holography and three-dimensional display: Principles
  and Applications}.\hskip 1em plus 0.5em minus 0.4em\relax Springer Science \&
  Business Media, 2006.

\bibitem{kim2010principles}
M.~K. Kim, ``Principles and techniques of digital holographic microscopy,''
  \emph{Journal of Photonics for Energy}, pp. 018\,005--018\,005, 2010.

\bibitem{nakatsuji2008free}
T.~Nakatsuji and K.~Matsushima, ``Free-viewpoint images captured using
  phase-shifting synthetic aperture digital holography,'' \emph{Applied
  optics}, vol.~47, no.~19, pp. D136--D143, 2008.

\bibitem{goodman2005introduction}
J.~W. Goodman, \emph{Introduction to Fourier optics}.\hskip 1em plus 0.5em
  minus 0.4em\relax Roberts and Company Publishers, 2005.

\bibitem{han2010laplacian}
Y.~Han and Q.~Yue, ``Laplacian differential reconstruction of one in-line
  digital hologram,'' \emph{Optics Communications}, vol. 283, no.~6, pp.
  929--931, 2010.

\bibitem{ren2015autofocusing}
Z.~Ren, N.~Chen, A.~Chan, and E.~Y. Lam, ``Autofocusing of optical scanning
  holography based on entropy minimization,'' in \emph{Digital Holography and
  Three-Dimensional Imaging}.\hskip 1em plus 0.5em minus 0.4em\relax Optical
  Society of America, 2015, pp. DT4A--4.

\bibitem{jiao2017enhanced}
S.~JIAO, P.~W.~M. Tsang, T.-C. Poon, J.-P. Liu, W.~Zou, and X.~Li, ``Enhanced
  autofocusing in optical scanning holography based on hologram
  decomposition,'' \emph{IEEE Transactions on Industrial Informatics}, 2017.

\bibitem{langehanenberg2008autofocusing}
P.~Langehanenberg, B.~Kemper, D.~Dirksen, and G.~Von~Bally, ``Autofocusing in
  digital holographic phase contrast microscopy on pure phase objects for live
  cell imaging,'' \emph{Applied optics}, vol.~47, no.~19, pp. D176--D182, 2008.

\bibitem{liebling2004autofocus}
M.~Liebling and M.~Unser, ``Autofocus for digital fresnel holograms by use of a
  fresnelet-sparsity criterion,'' \emph{JOSA A}, vol.~21, no.~12, pp.
  2424--2430, 2004.

\bibitem{memmolo2011automatic}
P.~Memmolo, C.~Distante, M.~Paturzo, A.~Finizio, P.~Ferraro, and B.~Javidi,
  ``Automatic focusing in digital holography and its application to stretched
  holograms,'' \emph{Optics letters}, vol.~36, no.~10, pp. 1945--1947, 2011.

\bibitem{goodfellow2016deep}
I.~Goodfellow, Y.~Bengio, and A.~Courville, \emph{Deep learning}.\hskip 1em
  plus 0.5em minus 0.4em\relax MIT press, 2016.

\bibitem{shimobaba2017convolutional}
T.~Shimobaba, N.~Kuwata, M.~Homma, T.~Takahashi, Y.~Nagahama, M.~Sano,
  S.~Hasegawa, R.~Hirayama, T.~Kakue, A.~Shiraki \emph{et~al.}, ``Convolutional
  neural network-based data page classification for holographic memory,''
  \emph{Applied optics}, vol.~56, no.~26, pp. 7327--7330, 2017.

\bibitem{muramatsu2017deepholo}
N.~Muramatsu, C.~W. Ooi, Y.~Itoh, and Y.~Ochiai, ``Deepholo: recognizing 3d
  objects using a binary-weighted computer-generated hologram,'' in
  \emph{SIGGRAPH Asia 2017 Posters}.\hskip 1em plus 0.5em minus 0.4em\relax
  ACM, 2017, p.~29.

\bibitem{kim2017deep}
S.-J. Kim, B.~Zhao, H.~Im, J.~Min, N.~Choi, C.~M. Castro, R.~Weissleder,
  H.~Lee, and K.~Lee, ``Deep-learning based hologram classification for
  molecular diagnostics,'' \emph{bioRxiv}, p. 192559, 2017.

\bibitem{rivenson2017phase}
Y.~Rivenson, Y.~Zhang, H.~Gunaydin, D.~Teng, and A.~Ozcan, ``Phase recovery and
  holographic image reconstruction using deep learning in neural networks,''
  \emph{arXiv preprint arXiv:1705.04286}, 2017.

\bibitem{shimobaba2017autoencoder}
T.~Shimobaba, Y.~Endo, R.~Hirayama, Y.~Nagahama, T.~Takahashi, T.~Nishitsuji,
  T.~Kakue, A.~Shiraki, N.~Takada, N.~Masuda \emph{et~al.}, ``Autoencoder-based
  holographic image restoration,'' \emph{Applied Optics}, vol.~56, no.~13, pp.
  F27--F30, 2017.

\bibitem{pitkaaho2017performance}
T.~Pitk{\"a}aho, A.~Manninen, and T.~J. Naughton, ``Performance of autofocus
  capability of deep convolutional neural networks in digital holographic
  microscopy,'' in \emph{Digital Holography and Three-Dimensional
  Imaging}.\hskip 1em plus 0.5em minus 0.4em\relax Optical Society of America,
  2017, pp. W2A--5.

\bibitem{pitkaaho2017focus}
T.~Pitk{\"a}aho, A.~Manninen, and T.~J. Naughton, ``Focus classification in digital holographic microscopy using deep
  convolutional neural networks,'' in \emph{European Conference on Biomedical
  Optics}.\hskip 1em plus 0.5em minus 0.4em\relax Optical Society of America,
  2017, p. 104140K.

\bibitem{griffin2007caltech}
G.~Griffin, A.~Holub, and P.~Perona, ``Caltech-256 object category dataset,''
  2007.

\bibitem{shimobaba2012computational}
T.~Shimobaba, J.~Weng, T.~Sakurai, N.~Okada, T.~Nishitsuji, N.~Takada,
  A.~Shiraki, N.~Masuda, and T.~Ito, ``Computational wave optics library for
  c++: {CWO++} library,'' \emph{Comput. Phys. Commun.}, vol. 183, no.~5, pp.
  1124--1138, 2012.

\end{thebibliography}

\end{document}